\documentclass{article}

\PassOptionsToPackage{numbers}{natbib}
\usepackage[final,dblblindworkshop]{neurips_2026}
\workshoptitle{Who Verifies the Agents?}
\makeatletter
\renewcommand{\@trackname}{\@neuripsordinal\ Conference on Neural Information Processing Systems (NeurIPS \@neuripsyear). Workshop: \@workshoptitle}
\makeatother

\usepackage[utf8]{inputenc}
\usepackage[T1]{fontenc}
\PassOptionsToPackage{hyphens}{url}
\usepackage{hyperref}

\usepackage{booktabs}
\usepackage{amsmath,amssymb,amsfonts}
\usepackage{nicefrac}
\usepackage{microtype}
\usepackage{graphicx}
\usepackage{multirow}
\usepackage{xcolor}
\usepackage{subcaption}
\usepackage{algorithm}
\usepackage{algpseudocode}
\usepackage{enumitem}
\usepackage{tikz}
\usetikzlibrary{arrows.meta,positioning,fit,backgrounds,calc}
\setlist{nosep,leftmargin=*}

\definecolor{tierAgreen}{HTML}{2C5F2D}
\definecolor{tierBorange}{HTML}{B85042}
\definecolor{tierCred}{HTML}{990011}
\definecolor{nexusblue}{HTML}{065A82}
\definecolor{sonnetC}{HTML}{1F6FB2}
\definecolor{opusC}{HTML}{D8590A}
\definecolor{gptC}{HTML}{2E8B57}
\definecolor{llamaC}{HTML}{9B5DE5}
\definecolor{decidT}{HTML}{E3EFE8}
\definecolor{judgT}{HTML}{F7E9DE}
\definecolor{contestT}{HTML}{E8E4EF}

\newsavebox{\routerA}\newsavebox{\routerB}\newlength{\routerH}

\title{Not Self-Decidable: LLMs Cannot Draw the Boundary\\of What an Agent Verifier Can Check}

\author{%
  Anthony Rhodes\\
  Confidential Core AI\\
  \texttt{a.rhodes@confidentialcore.ai}\\
}

\begin{document}
\maketitle

\begin{abstract}
A verifier for an agent faces rules of two kinds: the ones a fixed check can settle and the ones
that require a judge. A team that derives its own checks fixes that split up front. Where the
requirements come from outside, as they do in finance, healthcare and law, the agent enforces rules
it did not write, so the split falls to runtime, recurring for every predicate of every rule on
every action at a rate no reviewer can audit.

Every escalation scheme assumes a model can make that decision itself, that it is
\emph{self-decidable}. Across six corpora, including the EU AI Act, FINRA guidance and a deployed
credit agent, we collect roughly 22,000 labels from four models built by three labs. They agree
almost perfectly where the answer is obvious and collapse on regulatory text, where predicates pile
up on the boundary; their errors run in opposite directions, so no model can be trusted as the
conservative choice; and on the deployed agent's own rule-set they stop disagreeing and err
together, over-claiming that a fixed check will do, the direction that never gets escalated. We introduce
\textsc{CoVer} (corroborate-then-verify), an algorithm that treats unanimity as a nomination, admitting a
predicate only when the check synthesized for it survives intervention, reading fields the agent cannot write and holding under deterministic rewording. That gate rejects most of what corroboration wrongly admits, at a cost in coverage we report rather than tune away. Our central finding is that the obvious
alternative, agreement with a reference judge, certifies nothing: it climbs from 30\% to 77\%
across calibration bands while the genuinely decidable share does not move, because a judge drawn
from the population under indictment ratifies the blind spot it shares. Self-decidability is not a
capability to elicit from a model but a boundary the verifier must construct, and sharpening the instruments that construct it is where this problem becomes tractable.

\end{abstract}

\section{Introduction}
\label{sec:intro}

A verifier for an agent must know the limits of its own competence. Before it can check anything, something has to decide which of the rules it is asked to enforce can be settled by a fixed, interpretation-free procedure and which need a judge. The obligation ``logs are retained for twenty-four months'' is settled by a few lines of code; ``the disclosure is adequate'' is settled by no amount of code. Where the line between them falls is the \emph{decidability boundary}. In a deployed agent, that decision is not made once by a designer: it recurs for every predicate of every rule on every action, a predicate being one separate condition a rule imposes, with no human in the loop. Get it wrong in one direction and the system silently automates a judgment call; get it wrong in the other, and it spends a model call to re-derive what one line of code would settle. The obvious way to make the decision is to ask a model which rules are checkable. This paper is about why that does not work, and why it breaks down worst on exactly the operational rule-sets that agent systems enforce.

Using a model as a judge has become the default practice for evaluating and enforcing natural-language criteria, in reward modeling, benchmark scoring, safety filtering, agentic guardrails and automated policy compliance, on the implicit assumption that a capable enough model is a ``good-enough'' oracle for \emph{any} evaluation query. We term this the estimator-as-oracle default and argue in \S\ref{sec:axis} that it is a category error. Nearly all work on LLM judges asks whether the judge gets its answer right. We ask a prior question: whether a model can tell which questions need a judge at all. We call this \textbf{self-decidability}, and we show that models do not possess it on regulatory text, where the stakes are arguably highest for agentic security and compliance. 

Much of the value agentic systems promise is concentrated in regulated fields, where high-volume professional judgment meets stringent compliance obligations \cite{chen2024critical}, and where governing agents inside those obligations is already an active engineering problem \cite{raza2025trism}. An agent that extends credit, adjudicates a claim or drafts a disclosure, acts inside rules a regulator wrote; and compliance there is a property of each action rather than of the system in aggregate: a firm cannot certify that its agent is broadly suitable, only that \emph{this} recommendation, on \emph{this} account, satisfied the rule that governs it. Every obligation reaching the agent must therefore be settled somehow, and nobody upstream has sorted them first. In open-domain evaluation, the split is a design convenience and misclassifying a criterion costs a slightly noisy leaderboard. Under regulation, by contrast, the predicate set is \emph{externally imposed}, so the deploying party neither chooses which obligations it faces nor gets to rephrase the hard ones, and a misrouted judgment predicate means an unsupervised system certifies compliance it never assessed.

\begin{figure}[t]
\centering
\resizebox{\textwidth}{!}{%
\begin{tikzpicture}[
  font=\small,
  blob/.style  ={draw=black!52, line width=1.0pt},
  truth/.style ={draw=black!82, line width=1.2pt},
  edge/.style  ={line width=0.7pt},
  ttl/.style   ={font=\small\bfseries, anchor=west},
  lbl/.style   ={font=\scriptsize, text=black!70},
  sub/.style   ={font=\scriptsize\itshape, text=black!58},
  reg/.style   ={font=\scriptsize\bfseries, text=black!45},
  pin/.style   ={font=\scriptsize, text=black!72, fill=white, fill opacity=0.82,
                 text opacity=1, inner sep=1.1pt, rounded corners=0.5pt},
  halo/.style  ={draw=none, opacity=0.30},
]
\begin{scope}
  \node[ttl] at (0,56.12mm) {\textbf{A}\quad Decidability is a property of the criterion};
  \begin{scope}
    \clip (30.00mm,23.00mm) ellipse (28.20mm and 21.16mm);
    \fill[decidT] (-18.00mm,-13.80mm) rectangle (78.00mm,59.80mm);
    \fill[judgT] (30.00mm,-7.36mm) .. controls (35.10mm,11.96mm) and (29.70mm,28.52mm) .. (35.10mm,53.36mm) -- (78.00mm,53.36mm) -- (78.00mm,-7.36mm) -- cycle;
    \fill[halo, sonnetC!55] (12.30mm,23.00mm) ellipse (6.90mm and 11.27mm);
    \fill[halo, opusC!45]   (48.30mm,23.46mm) ellipse (6.00mm and 9.89mm);
    \fill[halo, llamaC!48]  (30.90mm,25.07mm) ellipse (7.50mm and 12.65mm);
    \fill[black!58] (9.00mm,21.16mm) circle (0.75pt);
    \fill[black!58] (11.40mm,28.52mm) circle (0.70pt);
    \fill[black!58] (13.80mm,16.56mm) circle (0.80pt);
    \fill[black!58] (10.20mm,33.12mm) circle (0.65pt);
    \fill[black!58] (15.60mm,24.84mm) circle (0.70pt);
    \fill[black!58] (12.60mm,12.88mm) circle (0.70pt);
    \fill[black!58] (16.80mm,30.36mm) circle (0.65pt);
    \fill[black!58] (8.40mm,25.76mm) circle (0.70pt);
    \fill[black!58] (15.00mm,20.70mm) circle (0.65pt);
    \fill[black!58] (12.00mm,18.40mm) circle (0.65pt);
    \fill[black!58] (47.40mm,21.16mm) circle (0.75pt);
    \fill[black!58] (50.40mm,26.68mm) circle (0.70pt);
    \fill[black!58] (45.60mm,15.64mm) circle (0.70pt);
    \fill[black!58] (51.60mm,22.08mm) circle (0.65pt);
    \fill[black!58] (48.60mm,31.28mm) circle (0.65pt);
    \fill[black!58] (43.80mm,25.30mm) circle (0.70pt);
    \fill[black!58] (50.40mm,17.48mm) circle (0.65pt);
    \fill[black!82] (25.80mm,15.64mm) circle (1.05pt);
    \fill[black!82] (28.20mm,23.00mm) circle (1.10pt);
    \fill[black!82] (30.60mm,29.44mm) circle (1.05pt);
    \fill[black!82] (33.00mm,19.32mm) circle (1.10pt);
    \fill[black!82] (34.80mm,26.68mm) circle (1.00pt);
    \fill[black!82] (27.60mm,33.12mm) circle (1.00pt);
    \fill[black!82] (36.00mm,22.08mm) circle (1.00pt);
    \fill[black!82] (24.60mm,26.68mm) circle (1.05pt);
    \fill[black!82] (31.80mm,14.26mm) circle (1.05pt);
    \fill[black!82] (29.40mm,17.94mm) circle (1.00pt);
    \fill[black!82] (33.60mm,32.20mm) circle (1.00pt);
    \fill[black!82] (26.40mm,30.36mm) circle (1.05pt);
    \fill[black!82] (36.60mm,17.48mm) circle (0.95pt);
    \fill[black!82] (30.00mm,35.88mm) circle (0.95pt);
    \fill[black!82] (34.20mm,23.92mm) circle (1.05pt);
    \fill[black!82] (27.00mm,19.78mm) circle (1.00pt);
    \fill[black!82] (35.40mm,28.98mm) circle (0.95pt);
    \fill[black!82] (28.20mm,26.22mm) circle (1.10pt);
    \fill[black!82] (31.20mm,21.16mm) circle (1.05pt);
    \fill[black!82] (32.40mm,16.56mm) circle (0.95pt);
    \draw[truth] (30.00mm,-7.36mm) .. controls (35.10mm,11.96mm) and (29.70mm,28.52mm) .. (35.10mm,53.36mm);
  \end{scope}
  \draw[blob] (30.00mm,23.00mm) ellipse (28.20mm and 21.16mm);
  \node[reg] at (12.00mm,41.40mm) {DECIDABLE};
  \node[reg] at (48.00mm,40.48mm) {JUDGMENT};
  \node[pin] at (11.10mm,7.13mm) {IFEval};
  \node[pin] at (48.90mm,8.05mm) {MT-Bench};
  \node[pin, align=center] at (30.60mm,3.20mm) {EU AI Act, FINRA,\\[-1.5pt] agent rules};
  \node[sub, anchor=north, align=center] at (30.00mm,-6.44mm)
    {Clear cases sit far from the line.\\[-1.5pt] Real governing text piles up on it.};
\end{scope}
\begin{scope}[xshift=74.00mm]
  \node[ttl] at (0,56.12mm) {\textbf{B}\quad Which model is the cautious one depends on the corpus};
  \node[lbl, anchor=south] at (22.00mm,45.60mm) {regulatory text};
  \fill[llamaC, opacity=0.15] (22.00mm,24.00mm) circle (17.00mm);
  \draw[edge, llamaC]         (22.00mm,24.00mm) circle (17.00mm);
  \fill[gptC, opacity=0.22]   (12.50mm,24.00mm) circle (9.04mm);
  \draw[edge, gptC]           (12.50mm,24.00mm) circle (9.04mm);
  \node[pin, text=llamaC!92, anchor=west]  at (27.60mm,33.20mm) {Llama};
  \node[pin, text=gptC!92]                 at (12.50mm,24.00mm) {GPT-4o};
    \node[sub, anchor=north, align=center] at (22.00mm,4.60mm)
    {GPT-4o's decidable set sits\\[-1.5pt] \emph{inside} Llama's};
  \node[lbl, anchor=south] at (72.00mm,45.60mm) {agent rules};
  \fill[gptC, opacity=0.15]   (72.00mm,24.00mm) circle (17.00mm);
  \draw[edge, gptC]           (72.00mm,24.00mm) circle (17.00mm);
  \fill[llamaC, opacity=0.22] (75.30mm,24.00mm) circle (14.96mm);
  \draw[edge, llamaC]         (75.30mm,24.00mm) circle (14.96mm);
  \node[pin, text=gptC!92, anchor=east]    at (60.20mm,35.60mm) {GPT-4o};
  \node[pin, text=llamaC!92]               at (76.60mm,24.00mm) {Llama};
    \node[sub, anchor=north, align=center] at (72.00mm,4.60mm)
    {the containment reverses:\\[-1.5pt] Llama's now sits inside GPT-4o's};
  \node[sub, anchor=north, align=center] at (47.00mm,-3.60mm)
    {The regions overlap but do not \emph{nest}: no model is uniformly conservative.};
  \path (-3.00mm,15.00mm) (100.00mm,15.00mm);
\end{scope}
\end{tikzpicture}}
\caption{\textbf{The geometry of the failure}. \textbf{A}~Admitting a fixed procedure is a property of the criterion (schematic): the calibration endpoints sit far from the line, and predicates decomposed from regulatory text pile up on it. \textbf{B}~Measured, not schematic. On the 231 regulatory-text predicates GPT-4o calls 41 decidable against Llama's 145, its set lying inside Llama's on all but 2; on the 73 agent rules the containment reverses, Llama's 48 inside GPT-4o's 62 on all but 1. The contested zone this leaves is 61\% of our regulatory-text predicates, while the endpoints stay unanimous ($\kappa$ 0.92--0.99).}
\label{fig:overview}
\end{figure}

On unambiguous cases, four foundation models spanning three labs agree almost perfectly on the decidable/judgment label ($\kappa$ 0.92--0.99): models \emph{can} draw the line when it is clear. However, on predicates decomposed from regulatory text, agreement collapses ($\kappa$ 0.20--0.50), and the failure is bidirectional: two models lean toward over-claiming decidability and two toward under-claiming, the split running across labs and within one, with a model's direction flipping by corpus. This indicates that there is no common bias to subtract out, no reliably conservative model to pick, and no lab to trust. The natural defense is that this is appropriate humility on a hard boundary, but the labels are confident rather than hedged, opposed rather than converging, localised to the operationally critical middle, and systematically unstable under meaning-preserving paraphrase of the \emph{same} predicate (\S\ref{sec:borderline}). A system can be humble about hard answers while remaining reliable about which questions are hard; conflating the two is fatal.

This matters because in industry today, the standard remedy for an unreliable judge is to simply escalate the hard cases \cite{moslem2026routing,zellinger2025abstention,rubriccond2026}. Every such remedy presupposes that the escalation decision itself is reliable. In this paper, we show that this presupposition does not hold, and the error is directional in the way that is systematically injurious: an over-claimed predicate is never escalated, so the mitigation is silently switched off exactly where it was needed. Whether that decision is reliable is a precondition for every escalation-based remedy, and to our knowledge no one has measured it heretofore. 

The decidability boundary therefore has to be established without trusting any single model's answer (Fig.~\ref{fig:overview}, \S\ref{sec:resolution}). What we build instead, \textsc{CoVer} (COrroborate-then-VERify predicate routing), has two parts. First, we ask several independent models and keep only the rules they all call checkable. That agreement is a cheap filter and a revealing one: it is 97--99\% precise where the models err in different directions, and only 83\% on the agent's own rules, where they err together. Filtering alone therefore cannot carry the guarantee. So, second, do not trust the label at all: write the check the rule implies and admit the rule only if that check survives intervention, reading fields the agent cannot write, moving when we intervene on them, and holding when the record is deterministically reworded. We built the more obvious second stage first, agreement with a reference judge, and \S\ref{sec:agreegate} shows why it cannot work. Correctness then rests on a check that was executed rather than a label that was asserted. In total, we provide the following contributions:

\begin{enumerate}
\item \textbf{The problem, named and located.} We identify \emph{self-decidability}, whether a model can tell which of the criteria it is given a fixed check can settle and which need a judge, as the routing decision every agent verifier makes before it checks anything. It is prior to the accuracy question the LLM-as-judge literature studies, and every escalation-based remedy silently assumes it has been solved -- it has not.
\item \textbf{\textsc{CoVer}}, a corroborate-then-verify procedure that keeps the model's answer off the critical path. Agreement across independent models only nominates a rule; admitting it requires a test the panel does not control. We show that the obvious such test, agreement with a reference judge, measures whether a check is \emph{reproducible} rather than whether a predicate is \emph{decidable}, and we replace it with an interventional gate that consults no estimator: the check must read fields the agent cannot write, respond to intervention on them, and hold under deterministic rewording of the record.
\item \textbf{The measurements.} 832 predicates from six corpora, each labelled by four foundation models across three labs under four rubric variants and two perturbation instruments: roughly 22,000 model calls. Agreement is near-perfect on unambiguous cases and collapses on regulatory text; the error is bidirectional, model-specific, cross-lab and domain-specific; the contested zone is the central mass of regulatory text rather than a tail; and a pre-registered ablation shows that a far more prescriptive rubric buys agreement without buying correctness.
\end{enumerate}

\section{Related Work}
\label{sec:related}

This is not another study of LLM-as-judge unreliability; surveys already document judges that are noisy, position- and verbosity-biased and domain-sensitive \cite{judgesurvey2024,li2024judgebias}. A more layered reading holds that these predicates are simply ambiguous and that human annotators would disagree too \cite{pavlick2019disagreement}; we answer it in Appendix~\ref{app:deferred}.

\textbf{Decidable-vs-subjective evaluation design.} Closest to us is Zhou and Shbita's \emph{Evaluating Ill-Defined Tasks in Large Language Models} \cite{zhou2026illdefined}, which characterises ill-defined tasks as those no fixed standard settles and routes decidable properties to rule-based checks, subjective ones to a judge. We share the distinction and differ on \emph{who draws the line, and where}: theirs is a guideline applied by hand, per benchmark, at construction time, while we study the regime with no designer in the loop, where the line falls to runtime and that automation is what fails. Where they use two judges to reduce noise \emph{within} the subjective region, we use corroboration to \emph{locate the boundary}, and then verify it.

A generic unreliability account makes three predictions, and the sections that follow falsify all three. It predicts broad degradation, where we measure agreement at $\kappa$ 0.92--0.99 on the endpoints, collapsing only in the consequential middle. It predicts undirected error, where we find the bias anti-correlated across models on regulation and \emph{correlated} on agent rules, and noise produces neither. And if the disagreement were underspecification, specifying the task should move the panel toward truth, whereas a decision procedure with worked examples moves it to 57.7\% accuracy against 88.8\% unprompted (\S\ref{sec:ablation}).


\section{The Setting: Runtime Verification of Agent Actions}
\label{sec:scale}
A single holistic evaluation, ``is this whole response good?'', can absorb a fuzzy estimator: one query, one judgment, and errors that wash out in aggregate scoring. By contrast, agentic deployment presents a very different error surface. An agent takes many actions per task, each is checked against the rules that apply, each rule breaks into predicates, and every predicate forces the same choice: (i) settle it with code, or (ii) spend a call on a judge. There is no third option, since a predicate nobody routes is still routed, by default, into one of the two. The multiplicities are large: RegData counts over a million binding restrictions in the US Code of Federal Regulations, marked by \emph{shall} and \emph{must} \cite{alubaydli2017regdata}, and one firm's subset runs to thousands. At that scale an executed check costs microseconds, where a judge costs seconds and tokens.

Critically, the two defaults cost different things (Figure~\ref{fig:dualcost}, Appendix~\ref{app:dualcost}). Take a firm required to disclose material conflicts of interest within three business days. Whether a conflict is \emph{material} is a judgment call, and sending it to code clears every disclosure that mentions a conflict at all, certifying compliance no one assessed. Whether it went out in three days is arithmetic on two timestamps, and sending it to a judge pays a model call to compare two dates. We call the first error \emph{over-claiming} and the second \emph{under-claiming}: over-claiming costs a missed violation, silent and repeated across a fleet, while under-claiming costs an extraneous inference one call at a time, which is how a year's inference budget disappears in
four months \cite{msv2026uber}.

We show that no deployment escapes both errors by simply \emph{choosing} a model, since the bias runs in opposite directions in different models (\S\ref{sec:results}): whichever minimises missed violations maximises the bill. Nor do the errors merely add up. A rule passes only if all of its predicates are handled correctly \emph{together}, so the chance every routing call is right degrades geometrically as rules lengthen. Appendix~\ref{app:dualcost} gives the cost model.

\section{The Decidability Axis, and Why Routing Needs Type}
\label{sec:axis}

\subsection{The boundary is not ours}
\label{sec:convergence}

Prior work in four fields independently discovered this decidability boundary in different guises, which is the best evidence available that it is a real seam in the criteria themselves rather than an artifact of our framing. Law separates \emph{rules}, whose content is fixed before anyone acts, from \emph{standards}, which acquire content only once conduct is judged, and Kaplow's tradeoff is the one we measure \cite{kaplow1992rules,hart1961}; psychology separates actuarial from clinical prediction \cite{dawes1989clinical}; psychiatry tried to move criteria \emph{across} the line by writing more specific ones and bought agreement without validity, the outcome we reproduce in \S\ref{sec:ablation} \cite{regier2013dsm5}; and reinforcement learning trains against tasks a program can check \cite{writingzero2025}. The nearest thing to a shared name for the distinction is Simon's between well-structured and ill-structured problems \cite{simon1973}.

What they share is that \emph{somebody drew the line in advance}, offline, once, by people with time and resources to adjudicate. Modern agentic deployment silently removes that assumption: an agent enforcing regulation it did not author
meets predicates nobody has pre-sorted, at a rate no designer can service. The boundary is therefore old; having to draw it at runtime,
without ground truth and without a person in the loop, is new.

\textbf{The estimator-as-oracle default.} An LLM judge is a stochastic estimator, not an oracle, and the criteria it is pointed at are non-homogeneous. Some are decidable: a fixed, interpretation-free procedure over stated facts returns the exact answer, whether a parser, a format or count check, a lookup, or an exact match. Others require judgment, turning on interpretation, degree or value, where estimation is the best available tool.\footnote{We use ``decidable'' by analogy with computability, not as a formal claim: a judgment criterion is not formally adjudicable or is underspecified, not Turing-undecidable.} Pointing one instrument at both wastes it where a check would suffice and over-trusts it where verification is required. The distinction is not about difficulty: a decidable predicate can be tedious to check, and a judgment predicate can be obvious to any person who reads it.

\emph{Type} -- not the verdict -- is what routing needs.
\label{sec:type}
Two questions can be asked about any criterion: (i) \emph{What is the verdict here?} is instance-specific and legitimately uncertain in hard cases; (ii) \emph{What kind of criterion is this?} asks whether settling it needs interpretation or only a check, and is a \emph{property of the criterion, not the case}. The latter question is what a routing system has to answer, which makes type what a verifier must be able to report. As we show, models turn out to be unreliable even about this (\S\ref{sec:stability}), which is why the architecture of \S\ref{sec:resolution} never depends on what they report.

\section{Experimental Setup}
\label{sec:setup}

Our experiments ask one question repeatedly: shown a single predicate, can a model say which side of
the decidability boundary it falls on? The corpora bracket that boundary, so a failure in the middle
is attributable to the boundary rather than to the task.
\textbf{Protocol.} We ask each model, per predicate, a single blind classification against a fixed
rubric, \emph{decidable = a fixed procedure over explicitly-stated facts returns the exact answer
with no interpretation, degree-weighing, or value judgment; otherwise judgment}, at temperature 0
with no hedging permitted.

\textbf{Scope.} The measurements rest on 832 distinct predicates from six corpora, each labelled independently by four foundation models spanning three labs, under four rubric variants and two perturbation instruments: approximately 22,000 model calls. Every figure and table below is computed directly from those label files.

\textbf{Corpora and models.} The four models are Claude Sonnet and Claude Opus \cite{anthropic2025claude}, GPT-4o \cite{openai2024gpt4o} and Llama-3.1-70B \cite{touvron2023llama}, all labelling every target; the cross-lab span is what lets the model-specificity claim rest on three cross-lab pairs.  \emph{Calibration endpoints}, where the label is uncontested, are IFEval \cite{ifeval2023} at the decidable pole and MT-Bench \cite{zheng2023judging} at the judgment one. The \emph{borderline} is policy: predicates decomposed from the EU AI Act \cite{euaiact2024} and FINRA \cite{finra2024ai}, \emph{regulatory text} throughout, and from an operational AI-agent rule-set, with a gold-labelled synthetic corpus as a control. Predicates were decomposed from rule text by an LLM under a fixed schema, blind to the labelling studied here; context-stripping would push the panel toward \textsc{judgment} \emph{together}, not into the bidirectional split we observe. Ground truth on the real corpora is an LLM-assigned proxy (\S\ref{sec:limitations}). Full protocol, metric definitions and model versions are in Appendix~\ref{app:extended}.

\begin{figure}[t]
\centering
\resizebox{\textwidth}{!}{%
\begin{tikzpicture}[
  font=\small,
  pnl/.style ={font=\small\bfseries},
  mlab/.style={font=\small\ttfamily, anchor=east},
  vin/.style ={font=\scriptsize, anchor=west, text=white},
  vinr/.style={font=\scriptsize, anchor=east, text=white},
  vout/.style={font=\scriptsize, text=black!75},
  tick/.style={font=\scriptsize, text=black!55},
  axis/.style={draw=black!65, thin},
  grid/.style={draw=black!12, very thin},
  over/.style ={fill=tierCred!75},
  under/.style={fill=tierBorange!45},
]
  \node[pnl] at (0mm, 11mm) {predicates from regulatory text};
  \draw[axis] (0mm,7mm) -- (0mm,-25mm);
  \draw[grid] (11.25mm,6mm) -- (11.25mm,-25mm);
  \node[tick] at (11.25mm,-27.5mm) {25\%};
  \draw[grid] (-11.25mm,6mm) -- (-11.25mm,-25mm);
  \node[tick] at (-11.25mm,-27.5mm) {25\%};
  \draw[grid] (22.5mm,6mm) -- (22.5mm,-25mm);
  \node[tick] at (22.5mm,-27.5mm) {50\%};
  \draw[grid] (-22.5mm,6mm) -- (-22.5mm,-25mm);
  \node[tick] at (-22.5mm,-27.5mm) {50\%};
  \draw[grid] (33.75mm,6mm) -- (33.75mm,-25mm);
  \node[tick] at (33.75mm,-27.5mm) {75\%};
  \draw[grid] (-33.75mm,6mm) -- (-33.75mm,-25mm);
  \node[tick] at (-33.75mm,-27.5mm) {75\%};
  \node[tick] at (0mm,-27.5mm) {0\%};
  \node[mlab] at (-40mm,0mm) {\texttt{sonnet}};
  \fill[under] (0mm,-2mm) rectangle (-5.4mm,2mm);
  \fill[over]  (0mm,-2mm) rectangle (9.9mm,2mm);
  \node[vout,anchor=east] at (-6.6000000000000005mm,0mm) {12\%};
  \node[vinr] at (8.4mm,0mm) {22\%};
  \node[mlab] at (-40mm,-6mm) {\texttt{opus}};
  \fill[under] (0mm,-8mm) rectangle (-24.3mm,-4mm);
  \fill[over]  (0mm,-8mm) rectangle (2.7mm,-4mm);
  \node[vin] at (-22.8mm,-6mm) {54\%};
  \node[vout,anchor=west] at (3.9000000000000004mm,-6mm) {6\%};
  \node[mlab] at (-40mm,-12mm) {\texttt{gpt4o}};
  \fill[under] (0mm,-14mm) rectangle (-33.300000000000004mm,-10mm);
  \fill[over]  (0mm,-14mm) rectangle (0.45mm,-10mm);
  \node[vin] at (-31.800000000000004mm,-12mm) {74\%};
  \node[vout,anchor=west] at (1.65mm,-12mm) {1\%};
  \node[mlab] at (-40mm,-18mm) {\texttt{llama70b}};
  \fill[under] (0mm,-20mm) rectangle (-10.8mm,-16mm);
  \fill[over]  (0mm,-20mm) rectangle (16.650000000000002mm,-16mm);
  \node[vin] at (-9.3mm,-18mm) {24\%};
  \node[vinr] at (15.150000000000002mm,-18mm) {37\%};
  \node[pnl] at (110mm, 11mm) {operational agent rules};
  \draw[axis] (110mm,7mm) -- (110mm,-25mm);
  \draw[grid] (121.25mm,6mm) -- (121.25mm,-25mm);
  \node[tick] at (121.25mm,-27.5mm) {25\%};
  \draw[grid] (98.75mm,6mm) -- (98.75mm,-25mm);
  \node[tick] at (98.75mm,-27.5mm) {25\%};
  \draw[grid] (132.5mm,6mm) -- (132.5mm,-25mm);
  \node[tick] at (132.5mm,-27.5mm) {50\%};
  \draw[grid] (87.5mm,6mm) -- (87.5mm,-25mm);
  \node[tick] at (87.5mm,-27.5mm) {50\%};
  \draw[grid] (143.75mm,6mm) -- (143.75mm,-25mm);
  \node[tick] at (143.75mm,-27.5mm) {75\%};
  \draw[grid] (76.25mm,6mm) -- (76.25mm,-25mm);
  \node[tick] at (76.25mm,-27.5mm) {75\%};
  \node[tick] at (110mm,-27.5mm) {0\%};
  \node[mlab] at (70mm,0mm) {\texttt{sonnet}};
  \fill[under] (110mm,-2mm) rectangle (106.4mm,2mm);
  \fill[over]  (110mm,-2mm) rectangle (137.9mm,2mm);
  \node[vout,anchor=east] at (105.2mm,0mm) {8\%};
  \node[vinr] at (136.4mm,0mm) {62\%};
  \node[mlab] at (70mm,-6mm) {\texttt{opus}};
  \fill[under] (110mm,-8mm) rectangle (104.6mm,-4mm);
  \fill[over]  (110mm,-8mm) rectangle (123.05mm,-4mm);
  \node[vout,anchor=east] at (103.39999999999999mm,-6mm) {12\%};
  \node[vinr] at (121.55mm,-6mm) {29\%};
  \node[mlab] at (70mm,-12mm) {\texttt{gpt4o}};
  \fill[under] (110mm,-14mm) rectangle (106.4mm,-10mm);
  \fill[over]  (110mm,-14mm) rectangle (141.95mm,-10mm);
  \node[vout,anchor=east] at (105.2mm,-12mm) {8\%};
  \node[vinr] at (140.45mm,-12mm) {71\%};
  \node[mlab] at (70mm,-18mm) {\texttt{llama70b}};
  \fill[under] (110mm,-20mm) rectangle (99.2mm,-16mm);
  \fill[over]  (110mm,-20mm) rectangle (130.7mm,-16mm);
  \node[vin] at (100.7mm,-18mm) {24\%};
  \node[vinr] at (129.2mm,-18mm) {46\%};
  \node[font=\scriptsize, text=tierBorange!90, anchor=east] at (52mm,-34mm)
    {$\leftarrow$ under-claim (wasteful): a check sent to a judge};
  \node[font=\scriptsize, text=tierCred!90, anchor=west] at (58mm,-34mm)
    {over-claim (unsafe): a judgment automated $\rightarrow$};
\end{tikzpicture}}
\caption{There is no conservative model to pick. On regulatory text (left) GPT-4o is the panel's most
cautious model, calling 1\% of judgment predicates decidable; on an operational agent rule-set
(right) the same model is the most reckless, at 71\%.
Every model shifts toward over-claiming on agent rules, the unsafe direction, and that is the
setting in which a runtime verifier acts on the label without escalation. Denominators differ: over-claim over
the true-judgment predicates (78 regulatory text, 24 agent rules), under-claim over the true-decidable
ones (153 and 49).}
\label{fig:flip}
\end{figure}

\section{Results}
\label{sec:results}
\label{sec:borderline}

\emph{Models can draw the line when it is clear.} On the calibration endpoints, IFEval and MT-Bench, all four models agree near-perfectly ($\kappa$ 0.92--0.99) with over- and under-claim near zero, the one exception being Opus over-claiming 10\% on clear judgment.

\emph{The boundary collapses on regulatory text.} Across the 231 predicates of regulatory text, agreement falls to $\kappa$ 0.20--0.50 under our rubric,\footnote{All $\kappa$ figures are measured under the single fixed rubric of \S\ref{sec:setup}. A far more prescriptive rubric raises regulatory-text $\kappa$ to 0.55--0.69, and \S\ref{sec:ablation} shows the extra agreement does not track ground truth; $\kappa$ is therefore not the rubric-invariant quantity here, and the precision of the unanimous set is. These marginals are also imbalanced, which depresses $\kappa$ by construction: four of six pairs reach 78--100\% of the $\kappa$ attainable at their own base rates. The models agree about as often as those rates permit; what differs is where each draws the line.} and the error is bidirectional, model-specific and cross-lab: Llama (37\%) and Sonnet (22\%) lean toward over-claiming while GPT-4o (74\%) and Opus (54\%) lean toward under-claiming, the split running \emph{within} Anthropic and \emph{across} labs, with the most opposed pair agreeing at only $\kappa$ 0.20. The lean is domain-specific too: on the operational agent rule-set every model flips to over-claiming (GPT-4o 71\%, Sonnet 62\%, Llama 46\%, Opus 29\%; $\kappa$ 0.38--0.64), so GPT-4o, the heaviest under-claimer on statutes, becomes the heaviest over-claimer on agent rules. A reasoning-era model added after these measurements extends the split rather than repairing it, over-claiming 41\% on regulatory text and 92\% on agent rules: the most permissive member of any panel we ran (Appendix~\ref{app:reasoning}). The flip is consequential: agent obligations \emph{read} mechanical, their conditions naming fields like \texttt{customer\_consent\_logged}, so models over-claim exactly where a system would auto-enforce.\footnote{Proxy caveat: ground truth here is the rule-set's \texttt{grading\_method}, and the 20 ``both'' rules count as judgment, which inflates absolute over-claim. We checked rather than assumed: recounting ``both'' as decidable leaves only four true-judgment predicates, and on those GPT-4o is still the heaviest over-claimer (75\%) and Opus the lightest (25\%). The ordering and the flip survive; the magnitudes rest on the ``both'' convention. $\kappa$ needs no ground truth and is unaffected.}

\emph{The contested zone is the central mass of regulatory text.} Measured ground-truth-free as the cross-model disagreement rate, the borderline covers 65\% of EU AI Act predicates, 52\% of FINRA and 33\% of agent rules, against 23\% synthetic. The rate rises monotonically with panel size (Table~\ref{tab:prevalence}), so these are lower bounds.

\textbf{The disagreement is a bug, not humility.} Genuine ambiguity predicts convergent hedging, spread evenly across the corpus and stable under rewording. We observe the opposite on each count. The labels are \emph{opposed} across models rather than converging, \emph{localised} to the operationally critical middle while the endpoints stay unanimous, and \emph{unstable} within a single model: under a loophole-free deterministic-template perturbation (\S\ref{sec:stability}) Opus flips its own borderline label 33\% of the time against 5\% on controls, and GPT-4o 44\% against 7\%. The last of these is instrument- and model-dependent, which we report as such, but every model destabilises under at least one perturbation. None of this depends on our proxy labels: on 53 FINRA rules a human annotator sorted blind, agreement is again $\kappa$ 0.34--0.48, with GPT-4o again the most conservative member (\S\ref{app:human}).

\emph{Corroboration recovers a high-precision core, and its failure mode is instructive.} Unanimous agreement yields a decidable set at 99.2\% precision on gold-labelled synthetic data and 97.0\% on regulatory text, but on the agentic corpus, where all four over-claim together, it falls to 83\%. That gap is the argument for \S\ref{sec:resolution}.

\section{Is the Boundary Just Our Rubric? A Specification Ablation}
\label{sec:ablation}

The obvious objection is that we measured ambiguity in \emph{our} prompt rather than unreliability in the models, so we red-teamed our own harness with two orthogonal manipulations of the published rubric, holding corpus, models and decoding fixed. \textbf{R-proc} restates the same criterion procedurally rather than definitionally, isolating \emph{surface form}; \textbf{R-spec} adds a decision procedure, boundary rules and worked examples, isolating \emph{underspecification}. We pre-committed to the bar that a fall of more than 5\,pp in contested rate on a real corpus would support the objection. Appendix~\ref{app:ablation} gives the design, table and pre-registration; four findings follow.

\emph{Surface form does not move the boundary.} Under R-proc, the contested rate is flat on every corpus while individual models flip 6.6--16.3\% of their own labels in no consistent direction (Appendix~\ref{app:ablation}).

\emph{Specification does move it, and our bar is crossed}: the contested rate falls 30.9\,pp on the EU AI Act and 19.6\,pp on FINRA. \emph{But the agreement it buys does not track truth.} Of the 71 regulatory-text predicates R-spec converts from contested to unanimous, the manufactured consensus is correct 57.7\% of the time (Wilson 95\%: 46.2--68.5) against 88.8\% (80.5--93.8) for consensus that formed unprompted, and the intervals do not overlap. That pits a hard subset against an easy one, so we also compare like for like: on those same 71 predicates the unprompted panel's own majority vote is right 49.3\% of the time, so specification does improve accuracy where it acts. What it does not do is beat a constant. The converted set is 74.6\% decidable, so a router that always said \textsc{decidable} would score 74.6\%, above R-spec and level with R-spec$'$. Twenty-nine of R-spec's 30 errors are under-claims: it does not resolve the boundary, it withdraws from it. Appendix~\ref{app:ablation} works one such predicate through in full.

\emph{And on operational agent rules it buys nothing} (33\% $\rightarrow$ 33\%, with half the
membership churning underneath).

\emph{Finally, which predicates it settles is an authoring accident.} R-spec$'$, identical to R-spec but for eight different and
equally correct worked examples, resolves a different third of the predicates (Jaccard 63.3\%) and is 75.9\%
accurate against R-spec's 57.7\% ($z=2.16$, $p=0.031$): the benefit of specification is a lottery over an authoring choice the rubric
does not record.

Specification is therefore a genuine lever on regulatory text but not on agent rules, and
where it works it suppresses disagreement rather than resolving it, moving the panel toward
each other and away from the answer. Contested rate is rubric-dependent; the precision of the
unanimous set is not, and that is what grounds \S\ref{sec:resolution}.

\section{Corroboration Nominates, Intervention Tests, Neither Certifies}
\label{sec:resolution}
Because no single model can be trusted to say whether a criterion admits a fixed procedure, we draw the boundary \emph{outside} any one estimator, in two layers: corroboration gives a high-precision candidate set cheaply, and verification audits that set with a test the panel does not control. \emph{Threat model.} We assume an agent that may be misaligned but is not attacking its verifier: predicates reach the router from a governance process rather than an adversary, and the fields a check reads are trusted in the restricted sense set out below. Adversarial authorship of the predicates themselves is out of scope (Appendix~\ref{app:deferred}).
\subsection{Corroboration as a model-agnostic safety floor}

\begin{table}[t]
\centering
\footnotesize
\setlength{\tabcolsep}{3.2pt}
\caption{The same routing question on both corpora. \textbf{Left}: no single model is
trustworthy in both directions, and the precise ones are precise by escalating most of the work;
unanimity reaches the high-precision corner without knowing which model is biased. \textbf{Right}: where the models err \emph{together} that precision collapses. Precision is against gold (syn.) and proxy labels; coverage is the share of true-decidable predicates recovered; false-decidables are gating-attributable; the agent corpus is 67.1\% decidable.}
\label{tab:routers}
\sbox{\routerA}{\begin{tabular}{@{}lccc@{}}
\toprule
\textbf{Router} & \textbf{Syn.} & \textbf{Proxy} & \textbf{Cov.} \\
\midrule
Sonnet          & 93.6\% & 88.8\% & 88.2\% \\
Opus            & 96.8\% & 93.3\% & 45.8\% \\
GPT-4o          & 99.2\% & 97.6\% & 26.1\% \\
Llama-3.1       & 86.4\% & 80.0\% & 75.8\% \\
Majority ($\ge$2/3) & 98.0\% & 93.7\% & 48.4\% \\
\textbf{Unanimous (4)} & \textbf{99.2\%} & \textbf{97.0\%} & 20.9\% \\
\bottomrule
\end{tabular}}%
\sbox{\routerB}{\begin{tabular}{@{}lccc@{}}
\toprule
\textbf{Router} & \textbf{Elim.} & \textbf{False-dec.} & \textbf{Prec.} \\
\midrule
Judge everything      &  0.0\% & 0 & -- \\
Majority ($\ge$3/4)   & 72.6\% & 9 & 83.0\% \\
Unanimous (4)         & 56.2\% & 7 & 82.9\% \\
\textbf{Intervention gate} & \textbf{27.4\%} & \textbf{3} & \textbf{85.0\%} \\
\bottomrule
\end{tabular}}%
\setlength{\routerH}{\dimexpr\ht\routerA+\dp\routerA\relax}%
\ifdim\dimexpr\ht\routerB+\dp\routerB\relax>\routerH
  \setlength{\routerH}{\dimexpr\ht\routerB+\dp\routerB\relax}\fi
\vspace{0.55\baselineskip}
\begin{minipage}[t]{0.475\textwidth}
\centering
\textit{regulatory text ($n=231$)}\\[-5pt]
\begin{minipage}[t][\routerH][c]{\linewidth}\centering\usebox{\routerA}\end{minipage}
\end{minipage}\hfill
\begin{minipage}[t]{0.505\textwidth}
\centering
\textit{agent rules ($n=73$)}\\[-5pt]
\begin{minipage}[t][\routerH][c]{\linewidth}\centering\usebox{\routerB}\end{minipage}
\end{minipage}
\end{table}

We keep only the predicates all four models independently call decidable (Table~\ref{tab:routers}).

\emph{The unanimous gate lands at the high-precision corner regardless of which model is biased,} bounding the safety-critical false-decidable rate to 1--3\%. It is also robust to panel membership, because the binding constraint is the most conservative member. We stated that prediction before testing it: adding a reasoning-era fifth member, which calls more regulatory predicates decidable than any model here, leaves the unanimous set \emph{identical} on both corpora (Appendix~\ref{app:reasoning}).

A skeptic will read the table as saying unanimity roughly equals the best single model, and ask why not use that one. \emph{Which} model is safe is domain-specific and not predictable in advance: ``just use GPT-4o'' presumes you know it is conservative on \emph{this} corpus, and on agent rules it is the heaviest over-claimer of the four. \emph{And precision is paid for in coverage} whichever way it is bought: GPT-4o routes three-quarters of the genuinely decidable to a judge and unanimity inherits that cost, reproducing the cost failure of \S\ref{sec:scale}.

\emph{And that precision is conditional on bias independence.} The 97--99\% figure holds only because the biases on regulation run in opposite directions. Where models share a \emph{directional} bias the panel agrees and is jointly wrong: on agent rules unanimous-\textsc{decidable} precision falls to 83\% at 69\% coverage. Corroboration is a cheap \emph{filter}, necessary but insufficient (\S\ref{sec:intervene}).

\subsection{The obvious second layer, and why it fails}
\label{sec:agreegate}

Since no panel can audit its own blind spot, the natural next move is to stop trusting the label and test it. We synthesize the check a predicate implies, run it against a reference judge on cases held out from the check's construction, and admit the predicate only when the two agree on at least a $\tau$ fraction of them. We ran that gate over EU AI Act predicates, using probe cases and reference verdicts collected before this study. We report it because the way it fails motivates what we adopt instead (Table~\ref{tab:cover}, Appendix~\ref{app:alg}).

At $\tau=1$, the gate admits 43/173 predicates at 79.1\% precision, which reads respectably until we ask where it comes from. Thirty-eight of the 173 have a
reference verdict that never varies, and the gate takes 16 of them, scoring 100.0\% there
against 66.7\% (Wilson 95\%: 47.8--81.4) on the 27 varying predicates it admits: a lift of 4.4 points over their 62.2\% base rate, with an interval that contains it. What looked like discrimination is composition.

The diagnostic underneath is sharper. Calibration agreement predicts held-out agreement almost perfectly, rising monotonically from 29.6\% to 76.7\% across the four buckets, so the gate measures something real. It is not measuring decidability: across those same buckets the true-decidable rate runs 55.6, 75.0, 62.8, 62.3 and 79.1\% and moves nowhere. We had built an instrument for reproducibility and read it as though it reported checkability.

The cause is structural, and it is the failure of \S\ref{sec:results} one level up: the reference judge is drawn from the same population of estimators as the panel, so wherever the panel shares a blind spot the judge shares it, and a check that reproduces the judge reproduces the blind spot. \textbf{Agreement with an estimator
cannot certify the absence of estimation.} No threshold repairs this, because $\tau$
governs only how faithfully the check tracks the judge, and tracking the judge is the
wrong target.

\subsection{Verification by intervention: \textsc{CoVer}'s second stage}
\label{sec:intervene}

Every failure we have measured is a failure of self-report: models disagree about what kind of criterion they face, disagree in opposite directions, contradict themselves under rewording, and a judge drawn from the same population inherits all of it. \textsc{CoVer}'s second stage therefore asks the models nothing further, admitting a predicate on
the behaviour of the synthesized check rather than on any estimator's report of it.

\begin{figure}[t]
\centering
\resizebox{\textwidth}{!}{%
\begin{tikzpicture}[
  font=\small,
  stage/.style ={draw=black!55, fill=black!3, rounded corners=1.3pt, line width=0.8pt,
                 minimum height=13mm, text width=25mm, align=center, inner sep=2.5pt},
  art/.style   ={draw=nexusblue!55, fill=nexusblue!6, rounded corners=1.3pt, line width=0.8pt,
                 minimum height=13mm, text width=40mm, align=left, inner sep=3pt},
  gate/.style  ={draw=black!55, fill=black!3, rounded corners=1.3pt, line width=0.8pt,
                 minimum height=13mm, text width=31mm, align=left, inner sep=3pt},
  admit/.style ={draw=tierAgreen!80, fill=tierAgreen!10, rounded corners=1.3pt,
                 line width=1.0pt, minimum height=13mm, text width=20mm,
                 align=center, inner sep=2.5pt},
  esc/.style   ={draw=tierBorange!80, fill=tierBorange!10, rounded corners=1.3pt,
                 line width=1.0pt, minimum height=8mm, text width=32mm,
                 align=center, inner sep=2.5pt},
  flow/.style  ={-{Straight Barb[length=1.8mm,width=1.6mm]}, line width=1.0pt, draw=nexusblue!85},
  drop/.style  ={-{Straight Barb[length=1.8mm,width=1.6mm]}, line width=0.9pt,
                 draw=tierBorange!85, densely dashed},
  lab/.style   ={font=\scriptsize, text=black!70, inner sep=1.5pt},
  labo/.style  ={font=\scriptsize, text=tierBorange!95, inner sep=1.5pt},
  tag/.style   ={font=\scriptsize\bfseries, text=black!55, inner sep=1pt},
]
\node[stage] (inp) at (0mm,0mm) {predicate $p$\\{\scriptsize ``retention $\ge$ 24 months''}};
\node[stage] (pan) at (36mm,0mm) {model panel $M$\\{\scriptsize all four call $p$}\\{\scriptsize \textsc{decidable}?}};
\node[art]   (syn) at (80mm,0mm) {\textsc{Synthesize}\\[1pt]
  {\scriptsize \textbf{slot} \texttt{retention\_months}}\\[-1pt]
  {\scriptsize \textbf{rule} \texttt{retention\_months >= 24}}};
\node[gate]  (gat) at (124mm,0mm) {{\scriptsize \textbf{G0} not an identity}\\[-1pt]
  {\scriptsize \textbf{G1} slots are attested}\\[-1pt]
  {\scriptsize \textbf{G2} verdict moves}\\[-1pt]
  {\scriptsize \textbf{G3} stable under rewording}};
\node[admit] (adm) at (159mm,0mm) {\textbf{admit}\\{\scriptsize $c$ enforces $p$}};
\draw[flow] (inp) -- (pan);
\draw[flow] (pan) -- (syn);
\draw[flow] (syn) -- (gat);
\draw[flow] (gat) -- (adm);
\node[tag, anchor=south] at (18mm,8.2mm)   {CORROBORATE};
\node[tag, anchor=south] at (102mm,8.2mm)  {VERIFY BY INTERVENTION};
\node[esc] (esc) at (80mm,-17mm) {\textbf{escalate}: $p$ is judged};
\draw[drop] (36mm,-7.5mm) -- (36mm,-17mm) -- (esc.west);
\draw[drop] (124mm,-7.5mm) -- (124mm,-17mm) -- (esc.east);
\node[labo, anchor=east, align=right] at (34mm,-11.5mm) {not\\unanimous};
\node[labo, anchor=west, align=left]  at (126mm,-11.5mm) {any gate\\fails};
\end{tikzpicture}}
\caption{\textsc{CoVer} end to end (Algorithm~\ref{alg:cv}). Corroboration only \emph{nominates}: a
predicate the panel does not unanimously call decidable leaves at once. A nominated predicate is
admitted only if the check synthesized for it clears all four gates. Both ways the router can err
lead out along the same dashed rail, so the gate fails closed, and no model is consulted after
nomination.}
\label{fig:cover}
\end{figure}

\textbf{What \textsc{Synthesize} emits.} A predicate becomes two artifacts (Fig.~\ref{fig:cover}): named fact-slots read off the action record, and a decision rule over them. A model fills the slots and code evaluates the rule. Only \emph{attested} fields are admissible, those infrastructure produces and the agent cannot write: a check over a field the agent \emph{asserts} establishes only that the agent agreed with itself, so a predicate needing one is escalated whatever the panel said. That distinction does its sharpest work where corroboration is most confidently wrong. An agent rule such as ``MUST NOT set \texttt{customer\_consent\_logged=True} without explicit consent'' reads mechanical because its field name does: it looks like a column, so the check looks like a lookup. But the agent writes that field, so a check reading it would certify the agent's own account of its compliance, and \textsc{CoVer} refuses the predicate corroboration most readily admits. If a record lacks a field the check needs, or the check raises an exception when it runs,
\textsc{CoVer} escalates the predicate rather than admitting it. The gate fails closed, so a broken
check costs a judge call and never a missed violation.

\textbf{Four gates, three inherited from the failures above.} \emph{Structure} rejects an identity that hands the predicate straight back to the extractor: the check-level form of the surface-mechanical reading that makes every model over-claim on agent rules. \emph{Groundedness} is the check-level form of specification that adds words without adding truth (\S\ref{sec:ablation}). \emph{Invariance} turns the perturbation instrument of \S\ref{sec:stability} on the check rather than the label. \emph{Sensitivity} is a degeneracy guard rather than a consequence of anything we measured, and we report it as such.

\textbf{Protocol.} We evaluate on the agent rule-set, where corroboration fails, running its 73 predicates against 36 synthetic action records (Appendix~\ref{app:alg}).

\emph{The gate carries real signal, and not enough of it.} Tightening invariance from half to full agreement lifts precision from 66.7\% to 85.0\% against a 67.1\% base rate, so requiring a check to survive rewording does select better predicates. On safety it does better than that column suggests: of the seven false-decidables that survive unanimity the gate rejects five, admitting 6.7 predicates per false-decidable against unanimity's 5.9. The operating point is chosen on a seeded half and confirmed on the other: there the gate admits 10 predicates at 90.0\% precision with one false-decidable, against unanimity's 19 at 89.5\% with two. What it does not do is clear our bar of beating unanimity's precision at comparable elimination, since it removes 27.4\% of judge calls where unanimity removes 56.2\%.

\emph{We can say why.} Our invariance test asks whether the extractor returns the same value when one record is rewrapped in neutral framing, never whether a slot means the same thing across records where the evidence is explicit and records where it is oblique. The gradient is a property \emph{between} records and the gate a statistic \emph{within} one, so it never reaches the gate. The instrument this calls for measures consistency across independent attempts rather than invariance under paraphrase. We built it. Sampling each extraction five times and scoring a predicate by its least stable read separates decidable from judgment at chance (AUC 0.42--0.48 on the 20 predicates that reached it, at every level from verdict to individual slot), and the median predicate of both classes is perfectly self-consistent. \texttt{not agent\_gave\_restructuring\_advice} reads identically every time: the extractor is not uncertain about the judgment it makes, it makes the same one. An estimator's agreement with itself certifies no more than its agreement with a judge.

\textbf{What intervention cannot do.} A check can be grounded, sensitive and invariant while verifying only part of what its predicate asks: one agent obligation carries three deontic clauses and its rule covers one. Establishing faithfulness would mean putting a judge back in. Each layer has one thing it cannot check: corroboration cannot certify itself, agreement cannot certify mechanicality, and intervention cannot certify faithfulness.

\section{Limitations and Conclusion}
\label{sec:limitations}
\label{sec:conclusion}

\textbf{Limitations.} \emph{Our agentic corpus is narrow}: 73 predicates over 20 rule families from one deployed credit agent, with the over-claim rates resting on its 24 true-judgment predicates, so Wilson 95\% intervals run roughly 35\,pp wide. The \emph{direction} is robust, not the magnitudes: every model shifts toward over-claiming, the largest shifts far outside sampling noise. Ground truth on the regulatory corpora is also an LLM-assigned proxy, mitigated by leading on the ground-truth-free disagreement metric and by replicating against 53 blind human labels (\S\ref{app:human}). Third, the interventional evaluation rests on 36 synthetic action records of our own construction, and its self-consistency null on the 20 predicates that reached that gate. Appendix~\ref{app:deferred} states six further limitations.

\textbf{Conclusion.} Pointing a single LLM judge at every evaluation treats a stochastic estimator as an oracle, and using one well means knowing \emph{which} criteria need it. That is what models cannot supply: they agree where the answer is obvious and err together where it matters, and no better judge repairs it, because the judge is what is being routed to. A verifier must know its own scope, and that knowledge is established by execution, not by asking
again. That requirement is sharpest under externally written rules: the EU AI Act, FINRA guidance, a bank's credit policy. There the routing error is silent, potentially catastrophic, and runs in the direction nobody escalates, and does not dissolve with scale, since the reasoning-era model we added widened the split rather than closing it. In regulated domains, where agentic systems carry the highest upside and the lowest tolerance for
error, the boundary of what a verifier can check must be settled in advance, not re-derived at every action from a model's
account of its own competence.

\bibliographystyle{plain}
\bibliography{references}

\clearpage\label{marker:appendixstart}\appendix

\section{The \textsc{CoVer} Procedure}
\label{app:alg}

Algorithm~\ref{alg:cv} states the procedure in full; the gates it applies are those of \S\ref{sec:intervene}.

\begin{algorithm}[htbp]
\caption{\textsc{CoVer}: corroborate-then-verify predicate routing. Corroboration orders the
synthesis budget rather than capping it: unanimous nominations are attempted first, and a
predicate the panel did not nominate can still be admitted if its check clears the gates.
No estimator's opinion enters lines 5--10.}
\label{alg:cv}
\begin{algorithmic}[1]
\Require predicates $P$; model panel $M$; action records $D_p$ for each $p$; deterministic
rewording operator $\rho$; grounding threshold $\theta_G$; invariance threshold $\theta_I$
\Ensure auto-enforced set $A$ of (predicate, check) pairs, and escalated set $E$
\State $A \gets \varnothing$;\quad $E \gets \varnothing$
\State $U \gets \{\, p \in P : \{\, m(p) : m \in M \,\} = \{\textsc{decidable}\} \,\}$
  \Comment{unanimous nominations: cheap, and not a verdict}
\For{each predicate $p \in U$, then each $p \in P \setminus U$ while budget remains}
  \State $c \gets \Call{Synthesize}{p}$ \Comment{bounded check: comparison, lookup, format or count}
  \If{$c$ reduces to an identity on one slot} \State $E \gets E \cup \{p\}$; \textbf{continue}
     \Comment{G0: structure}
  \EndIf
  \State $g \gets$ fraction of $c$'s slots readable from the attested fields of $D_p$
     \Comment{G1: groundedness}
  \State $s \gets \bigl[\, \exists\ \text{slot } \sigma,\ x \in D_p,\ \text{value } \sigma' :
     c(x) \neq c(x \,|\, \sigma \!\leftarrow\! \sigma') \,\bigr]$ \Comment{G2: sensitivity}
  \State $v \gets \bigl|\{\, x \in D_p : c(x) = c(\rho(x)) \,\}\bigr| \,/\, |D_p|$
     \Comment{G3: invariance}
  \If{$g \ge \theta_G$ \textbf{and} $s$ \textbf{and} $v \ge \theta_I$}
     \State $A \gets A \cup \{(p,c)\}$ \Comment{$c$ replaces the judge for $p$ at runtime}
  \Else
     \State $E \gets E \cup \{p\}$ \Comment{$p$ is judged, permanently}
  \EndIf
\EndFor
\State every $p$ never attempted joins $E$ \Comment{unattempted predicates are judged}
\State \Return $A,\ E$
\end{algorithmic}
\end{algorithm}

\begin{table}[htbp]
\centering
\small
\caption{The agreement gate on the EU AI Act corpus, against the label routers. Label rows
use all 175 labelled predicates, graduation rows the 173 the gate ran on. \emph{Held} is
agreement with the reference judge on held-out cases, which a label router cannot report
because a label is not executable.}
\label{tab:cover}
\setlength{\tabcolsep}{4pt}
\begin{tabular}{@{}lcccc@{}}
\toprule
\textbf{Router} & \textbf{Admitted} & \textbf{Prec.} & \textbf{Cov.} & \textbf{Held} \\
\midrule
Majority ($\ge$3 of 4)         &  56 & 92.9\% & 44.4\% & -- \\
Unanimous (4)                  &  24 & 95.8\% & 19.7\% & -- \\
Graduated ($\tau=1$)           &  43 & 79.1\% & 29.3\% & 76.7\% \\
Unanimous \emph{and} graduated &   7 & 100.0\% &  6.0\% & 76.2\% \\
\bottomrule
\end{tabular}
\end{table}

\subsection{Protocol for the interventional evaluation}

\textbf{Protocol.} We evaluate on the agent rule-set, where corroboration fails because the models err together. The 73 predicates run against 36 synthetic action records, each carrying a full session: tool-call log, session fields and conversation. Every judgment-bearing phenomenon appears in \emph{clear}, \emph{ambiguous} and \emph{absent} form, so a slot that can only be filled by weighing has somewhere to come apart, and the records state evidence but never a verdict. One record is shown to \textsc{Synthesize} as a schema exemplar and withheld from every gate. Because we inspected this corpus while building the harness, we fix a seeded split and report on the half that plays no part in choosing the operating point.

\textbf{The self-consistency instrument, in full.} \S\ref{sec:intervene} diagnoses our invariance
gate as a statistic \emph{within} a record where the difficulty gradient is a property \emph{between}
records, so we built the instrument that diagnosis calls for and report it here because it fails.
For each predicate we re-extract every slot $K=5$ times independently at temperature $1.0$ on a fixed
prefix of 12 records, and score each record by the share of attempts agreeing with the modal read.
The record set is bounded deliberately: a worst-case-over-records statistic on an unbounded set
rejects everything, because any predicate eventually meets one flaky read. We reduced these
measurements ten ways: at the verdict, at the full slot tuple and per individual slot, each under
worst-case, second-worst, first-quartile and mean. No aggregation choice can be blamed for the
outcome, because every one separates true-decidable from true-judgment predicates at chance
(AUC $0.42$--$0.48$; 13 decidable and 7 judgment predicates reached the gate), and 9 of 13 decidable
and 5 of 7 judgment predicates score perfect self-consistency. Two caveats bound this. The predicates
that reached the gate are not a random sample of the corpus, so the estimate is a bounded check
rather than a corpus-wide measurement; and constrained JSON extraction may simply be near-deterministic
at any temperature, in which case the instrument is inert rather than the hypothesis false. Neither
caveat touches the mechanism the result illustrates: \texttt{not agent\_gave\_restructuring\_advice}
delegates an entire judgment to the extractor and is read back identically on every record, because a
model that is confidently wrong is indistinguishable, to any reproducibility test, from one that is right.

\section{Perturbation Instruments for the Stability Test}
\label{app:stability}

The remaining feature-reading prediction is \emph{internal consistency}: even if models differ from each other, a single model should draw the same decidability line for a given predicate regardless of trivial surface wording, its line would be principled, just personal. We test this directly. For each contested predicate we generate $K$ neutral, meaning-preserving paraphrases, validate each is semantically equivalent to the original (an independent equivalence check, so we cannot be accused of adversarial rewording), and re-probe each model. We measure each model's flip rate: the fraction of predicates on which it contradicts \emph{itself} across paraphrases. As a control we run the same protocol on the unanimous-clear predicates.

\begin{itemize}
\item \emph{Bug prediction:} high flip rate on the borderline, $\approx$ 0 on the clear control, the instability is localized to exactly where the routing decision matters.
\item \emph{Feature prediction:} low flip rate everywhere (internal consistency).
\end{itemize}

\textbf{Result (two independent perturbation instruments).} We test internal consistency two ways: an LLM paraphrase instrument (Haiku paraphraser + Haiku equivalence gate, both outside the judge set; 134 borderline / 48 control parents) and a deterministic template instrument (neutral surface wrappers, no LLM in the loop, closing any ``the paraphraser drifted it / rubber-stamped it'' objection; 172 borderline / 60 control). Borderline flip = the model contradicts its \emph{own} label across meaning-preserving variants.

Three honest reads, stated plainly:

\begin{itemize}
\item \emph{The effect is real and, on the loophole-free instrument, large.} Opus (33\% vs 5\%) and GPT-4o (44\% vs 7\%) flip their own borderline label far above their clean-case baseline under purely deterministic reframing, with no LLM authoring or approving the variants. GPT-4o, nearly stable under paraphrase, is the \emph{most} unstable under templates: its apparent paraphrase-stability was an artifact of its near-constant-JUDGMENT under-claim bias, which the template wrappers break.
\item \emph{It is instrument- and model-dependent, and we do not hide that.} No single model destabilizes on every perturbation type: Sonnet is significant under paraphrase but flat under templates; Llama shows little within-model instability on either. The claim is therefore not ``every model is internally chaotic everywhere,'' but ``for any given model there exists a meaning-preserving perturbation under which its borderline self-classification is substantially noisier than its clean-case baseline'', which is already incompatible with the feature reading's prediction of a stable, principled internal line.
\item \emph{The control baseline is not zero (5--10\%).} The claim is excess borderline instability, not a perfect-versus-chaotic contrast.
\end{itemize}

Read together with \S\ref{sec:borderline}, the full picture is: the dominant, universal failure is \emph{cross-model}, all four models draw incompatible lines ($\kappa$ collapse to 0.20--0.50 on regulatory text, and to 0.41--0.71 on the full policy set, which includes the easier synthetic control; on the contested subset alone agreement is far lower), and the bias is bidirectional, model-specific, and cross-lab. The within-model instability is a secondary, corroborating signal that even a single model's line is not the principled invariant the feature reading requires. Every model is unreliable on at least one axis, Llama chiefly via cross-model bias, Opus/GPT-4o/Sonnet via within-model instability as well. Neither across models nor reliably within a single model does self-report yield a trustworthy routing signal, and that, applied to the contested mass that is half-to-two-thirds of regulatory text, is what makes single-estimator self-routing operationally disqualifying at scale (\S\ref{sec:scale}).

\section{Extended Protocol and Related Work}
\label{app:extended}

\subsection{Models and corpora}

\textbf{Models.} Four instruction-tuned foundation models span three labs: Claude Sonnet and Claude Opus \cite{anthropic2025claude}, GPT-4o \cite{openai2024gpt4o}, and Llama-3.1-70B \cite{touvron2023llama}. All four label every target. The cross-lab span is deliberate, since the model-specificity claim requires showing that disagreement is not an artifact of one lab's training, and the three cross-lab pairs carry that argument. Figure~\ref{fig:kappa} gives every pairwise agreement across three corpora.

\textbf{Data and ground truth.} \emph{Calibration endpoints}, where the label is essentially uncontested, are IFEval \cite{ifeval2023}, whose instructions are verifiable by construction, as the clearly-decidable pole and MT-Bench \cite{zheng2023judging} as the clearly-judgment one. The \emph{borderline} comes from policy: predicates decomposed from the EU AI Act \cite{euaiact2024} and FINRA \cite{finra2024ai}, and from an operational AI-agent rule-set, with a gold-labelled synthetic corpus as a controlled supplement rather than a headline. Ground truth on the two regulatory corpora is a proxy: an LLM assigned each predicate a grounding level from rule and predicate text alone, blind to the panel's outputs, under a fixed scheme in which the two grounded levels map to \textsc{decidable} and the two ungrounded ones to \textsc{judgment}. We therefore lead the vs-ground-truth claims on the gold-labelled synthetic corpus and the endpoints, and lead the headline claims on the ground-truth-free disagreement metric, which needs no labels at all. \S\ref{app:human} reports a replication against human labels.

\textbf{Reproducibility note.} The panel cannot be reconstructed exactly: the Sonnet checkpoint used for the labels reported here was retired by its provider after our run. Our released label files therefore constitute the primary artifact. The OpenAI model was specified by a floating alias rather than a dated snapshot, so its exact revision is not recoverable either. Where we re-probe (\S\ref{sec:stability}) we hold the panel fixed within a comparison.

\subsection{Replication against human labels}
\label{app:human}

Because the regulatory-corpus ground truth is LLM-assigned, we replicated the core measurement against labels a machine did not produce. A human annotator sorted 54 FINRA rules into three enforceability tiers, \textbf{A} settleable by a fixed interpretation-free procedure, \textbf{B} structured-ambiguous, and \textbf{C} requiring judgment, working from written tier definitions fixed before annotation and blind both to every model output and to the deterministic signal file. One rule was excluded, leaving 53. We map A to \textsc{decidable} and B and C to \textsc{judgment}, and additionally report the A-versus-C contrast with the structured-ambiguous middle removed. The rubric, decoding and panel are those of \S\ref{sec:setup}, with the retired Sonnet checkpoint replaced by its nearest dated successor.

\textbf{What holds.} Pairwise agreement is $\kappa$ 0.34--0.48, inside the 0.20--0.50 we report for regulatory text, and the models contest 34\% of the 53 rules (44.7\% with the middle tier removed) against 52\% on FINRA predicates. The ordering holds where it matters: GPT-4o is again the most conservative member, under-claiming 85\% of decidable rules while over-claiming 3\%, and the panel spreads by a factor of nearly four in how much it will automate at all, from GPT-4o's 4 rules to Opus's 15 of the same 53. Unanimity again buys precision at the cost of coverage: on the A-versus-C contrast the unanimous set is 100\% precise over 10\% of the decidable rules.

\textbf{What does not.} The error is \emph{not} bidirectional here. Every model leans the same way, over-claiming 3--18\% and under-claiming 45--85\%, where on decomposed regulatory predicates two models lean each way. We attribute this to granularity rather than to a failure of the earlier result: whole rules seldom admit a fixed procedure, and the human labels put only 20 of 53 on the decidable side against 153 of 231 for predicates. The consequence for \S\ref{sec:resolution} is worth stating, since the 97--99\% unanimous precision there is explained by bias independence: on a corpus where the panel leans one way, unanimity's precision is inherited from the panel's shared caution rather than from errors cancelling.

\textbf{Scope.} Fifty-three rules give intervals 15 to 30 points wide, so this corroborates a pattern and does not sharpen a number. The targets are whole rules rather than decomposed predicates, and the annotation is by one person. Independent multi-annotator labelling at predicate granularity remains the strongest available improvement to this evidence base.

\textbf{Protocol.} Each model labels each predicate, blind, as DECIDABLE or JUDGMENT against one fixed rubric, at temperature 0, with no hedging permitted. Labels are single-draw, and we verify that this is adequate rather than assuming it: re-running the identical rubric and corpus months later reproduced each unchanged model's own labels at 97.2\% (GPT-4o), 99.0\% (Opus) and 99.4\% (Llama-3.1). The one model whose labels did not reproduce had been version-upgraded in the interval, which is the subject of \S\ref{sec:ablation}. We compare each label to ground truth where available and to the other models. For the within-model stability test (\S\ref{sec:stability}) we re-probe contested predicates under meaning-preserving perturbations: an LLM-paraphrase instrument (paraphrases authored and equivalence-validated by a model outside the judge set) and a deterministic template instrument (neutral surface wrappers, no model in the loop), with a control group of unanimous-clear predicates.

\textbf{Metrics.} (i) Over-claim and under-claim rates vs.\ ground truth, kept on their separate denominators (\S\ref{sec:axis}). (ii) Cross-model agreement via Cohen's $\kappa$ \cite{cohen1960kappa}, on the endpoints and the borderline. (iii) The contested rate, fraction of predicates without a unanimous label, as a \emph{ground-truth-free} measure of borderline prevalence. (iv) Corroboration precision and coverage: among unanimous-DECIDABLE predicates, the fraction truly decidable, and the share of true-decidable recovered, versus single-model self-report. (v) Within-model flip rate on the borderline vs.\ control. For the enforcement capstone (Appendix~\ref{app:capstone}) we additionally report rule-level fidelity vs.\ a holistic reference, the deterministically-automated fraction, and the gating-attributable false-clear rate.

\textbf{Programmatic / verify-don't-judge evaluation.} Program-as-a-judge approaches (PAJAMA \cite{pajama2025} synthesize executable judging programs instead of scoring with an LLM, cutting cost and bias; reinforcement learning with verifiable rewards (RLVR; e.g. Writing-Zero \cite{writingzero2025} couples training to checkable signals and treats verifiability as given by domain (math/code checkable, open-ended not). Both \emph{assume} you know what is program-checkable. We address the prior question they skip: which predicates are checkable when the model cannot self-identify them, and how to establish that boundary safely (corroboration + a verification gate) rather than presupposing it.

\textbf{Metacognition and self-knowledge.} Work on whether models know what they know calibrates confidence in \emph{answers} \cite{kadavath2022know,mirror2026}. Our type-vs-answer distinction (\S\ref{sec:type}) is orthogonal: a model can be well calibrated about whether its \emph{answer} is right while being unreliable about whether the \emph{question} has a checkable answer at all.

\textbf{Agentic runtime enforcement (application context).} Neuro-symbolic oversight and runtime policy enforcement for agents (e.g. FormalJudge \cite{formaljudge2026}; AgentSpec \cite{wang2025agentspec}) presuppose a policy already partitioned into checkable vs.\ judged obligations; our contribution is upstream, establishing that partition reliably is a prerequisite for any such system.

\subsection{A reasoning-era panel member}
\label{app:reasoning}

The panel of \S\ref{sec:setup} predates wide availability of models trained to deliberate before
answering, and type-classification is plausibly the kind of task extended reasoning helps. We
therefore added a fifth labeller after the measurements above were complete and fixed: an
open-weight reasoning distillation (DeepSeek-R1-Distill-Qwen-14B, served locally, temperature 0,
2{,}000-token budget so the deliberation trace is never truncated). Traces are stripped before
parsing, and a response from which no label can be recovered is recorded as missing rather than
defaulted, so a truncated deliberation cannot masquerade as a \textsc{judgment} label. It labelled the 559 policy predicates, the 73 agent rules and the 200 calibration endpoints; it is not part of the four-model panel used everywhere else in this paper, and no figure or table above includes it.

\textbf{Deliberation does not repair the boundary; here it widens the damage.} On regulatory text
the model calls 73.2\% of predicates decidable, more than Sonnet (65.8\%), Llama-3.1 (62.8\%), Opus
(32.5\%) or GPT-4o (17.7\%), which makes it the most permissive labeller we measured. Its
over-claim rate is 47\% on EU AI Act judgment predicates and 25\% on FINRA. On the operational
agent rule-set it over-claims on 92\% of judgment predicates and calls 95.9\% of all predicates
decidable, so the domain flip of \S\ref{sec:borderline} holds for a fifth model and reaches its
largest value there. Pairwise $\kappa$ against the incumbents is 0.13 (GPT-4o), 0.26 (Opus), 0.54
(Sonnet) and 0.59 (Llama-3.1): adding a more deliberative model lowers the floor of the agreement
range rather than raising it.

\textbf{It is permissive at the judgment pole as well.} On IFEval the model is exactly right, calling all 120 instructions decidable. On MT-Bench it calls 30\% of them decidable, against 0.0\% for GPT-4o, 1.2\% for Sonnet and Llama-3.1 and 10.0\% for Opus, so its permissiveness is an offset present even where the incumbents are near-perfect rather than something the borderline induces. Where it errs is orderly: over-claiming runs at 50\% on \emph{math}, \emph{stem} and \emph{extraction}, 30\% on \emph{reasoning} and 20\% on \emph{roleplay}, and vanishes on \emph{humanities} and \emph{writing}. A model trained against verifiable rewards in mathematics and code appears to have learned that those domains admit checkable answers, and to carry the inference across to the claim that the \emph{quality of a response} in them is checkable too. That is the estimator-as-oracle error of \S\ref{sec:axis} reached by a training route rather than by a capability limit, which is a further reason not to expect scale alone to remove it.

\textbf{The unanimous gate is unmoved, as predicted.} \S\ref{sec:resolution} argues that
corroboration is bound by its most conservative member and is therefore insensitive to who else
joins. Adding the most permissive labeller in this paper leaves the unanimous-\textsc{decidable}
set unchanged on both corpora: 33 predicates at 97.0\% precision and 20.9\% coverage on regulatory
text, and 41 at 82.9\% with the same seven false-decidables on agent rules, in each case identical
to the four-model gate. The prediction was in the text before the experiment was run.

\textbf{Scope.} This is one open-weight distillation at 14B, not a frontier reasoning system, so it
speaks to whether deliberation repairs self-decidability and not to whether capability does. The
direction is nonetheless the one that matters: a model given time to think became the panel's
least safe member on the corpus where a verifier would act on its label without escalation.

\section{Specification Ablation in Full}
\label{app:ablation}

The obvious objection is that we measured ambiguity in \emph{our} classification prompt rather than
unreliability in the models. We red-teamed our own harness, and report the result including the
part that goes against us. Two orthogonal manipulations of the published rubric hold corpus, models
and decoding fixed: R-proc (202 tokens) restates the same criterion procedurally rather than
definitionally, isolating \emph{surface form}; R-spec (707 tokens) adds a four-step decision procedure,
five boundary rules and eight worked examples, isolating \emph{underspecification}. We pre-committed
to the bar that a fall of more than 5\,pp in contested rate on a real corpus would support the
objection.

\begin{table}[t]
\centering\small
\begin{tabular}{lrrrr}
\toprule
Corpus & $n$ & R-base & R-proc & R-spec \\
\midrule
EU AI Act        & 175 & 65\% & 59\% & \textbf{34\%} \\
FINRA            &  56 & 52\% & 55\% & \textbf{32\%} \\
Agent rule-set   &  73 & 33\% & 40\% & \textbf{33\%} \\
Synthetic        & 328 & 23\% & 23\% & 11\% \\
Endpoints        & 120 &  0\% &  0\% &  0\% \\
\bottomrule
\end{tabular}
\caption{Contested rate (no unanimous label across four models) by corpus and rubric.
R-base is the published rubric.}
\label{tab:ablation}
\end{table}

\textbf{Surface form does not move the boundary} (Table~\ref{tab:ablation}). Under R-proc the contested rate is flat on
every corpus ($-6$, $+3$, $+7$, $0$, $0$\,pp) while individual models flip 6.6--16.3\% of
their own labels, in no consistent direction against ground truth. Rewording changes which
predicates a model calls decidable without changing how much the panel disagrees.

\emph{Specification does move it, and our pre-committed bar is crossed.} Under R-spec the
contested rate falls 30.9\,pp on the EU AI Act and 19.6\,pp on FINRA, far beyond the 5\,pp
bar. On the contested-rate metric, the objection holds for both regulatory corpora, and we
report it as such.

\emph{But the agreement R-spec buys does not track truth.} Applying this paper's own corroboration
measure to the ablation: of the 71 regulatory-text predicates R-spec converts from contested to
unanimous, the manufactured consensus is correct 57.7\% of the time (Wilson 95\%: 46.2--68.5),
against 88.8\% (80.5--93.8) for consensus that formed under R-base without prompting, and 90.4\% for the
R-base-unanimous subset that survives into R-spec. The intervals do not overlap. One confound deserves naming, that if the proxy annotator shared the unprompted
models' intuitions then R-base-unanimous would agree with it partly by construction; the direction of
the errors rules this out, since R-spec's consensus errs 97\% toward JUDGMENT against a reference set
that is 74.6\% decidable, so the confound would require an annotator who systematically over-calls \emph{decidable}. The manufactured consensus is 35.2\% decidable against that 74.6\%. Twenty-nine of the 30 errors are under-claims. R-spec does not resolve the boundary,
it withdraws from it, routing to a judge what a program could settle: the expensive corner of the
dual-cost model rather than a solution to it, with safety preserved and the automation the routing
decision exists to enable forfeited.

\emph{On operational agent rules, specification buys nothing.} The contested rate is unchanged
(33\% $\rightarrow$ 33\%), and even that stability is coincidental: half the membership churns, 12
predicates becoming contested and 12 unanimous. The corpus where a runtime verifier actually acts
is the corpus a decision procedure with worked examples cannot help.

\emph{Swap the examples and the consensus moves.} We pre-registered four predictions and ran R-spec$'$,
identical to R-spec to the character but for the eight worked examples, replaced by eight different
correct ones. B1 confirmed: contested rate 40.7\%, inside the predicted 25--41\% band. B2 confirmed:
the predicates R-spec$'$ resolves overlap the R-spec-resolved set at Jaccard 63.3\%, so swapping the examples
changes roughly a third of which predicates get settled at all. B3 \emph{failed}: we predicted the
consensus would again fall below 70\% accurate, and it is 75.9\% (63.5--85.0). B4, the falsifier that
would have shown specification resolving a stable and correct subset, did not trigger. The failed prediction is the
more informative one. Two sets of worked examples, both correct and both authored in good faith,
inside an otherwise character-identical rubric, yield consensus that is 58\% and 76\% accurate: the
benefit of specification is a lottery over an authoring choice the rubric does not record. The
direction is unchanged, all 14 errors made by R-spec$'$ being under-claims, against 29 of R-spec's 30.

\textbf{What the ablation establishes.} The disagreement we measure is not an artifact of one
phrasing: a different phrasing produces different labels and the same disagreement, and a far more
prescriptive one produces agreement whose accuracy swings 18 points on which examples were written,
landing 13 to 31 points below consensus that formed unprompted. No rubric we wrote made the panel
both agree and be right, and none of them moved the accuracy of the agreement that formed without
prompting.


\section{Deferred Instruments and Diagnostics}
\label{app:deferred}

The following material expands \S\ref{sec:borderline}. The body states the headline numbers and
cites \S\ref{sec:stability} for the stability result; this appendix gives the instruments, the full
tables and the diagnostics behind them.

\begin{table}[h]
\centering
\small
\caption{Contested rate (no unanimous decidable/judgment label) by corpus and panel size. Measured ground-truth-free as cross-model disagreement.}
\label{tab:prevalence}
\begin{tabular}{@{}lcc@{}}
\toprule
\textbf{Corpus} & \textbf{3-model} & \textbf{4-model} \\
\midrule
EU AI Act (regulatory text) & 54\% & \textbf{65\%} \\
FINRA (regulatory text) & 43\% & \textbf{52\%} \\
Synthetic (designed, gold labels) & 16\% & 23\% \\
Agent rule-set (operational, agentic) & 22\% & \textbf{33\%} \\
Pooled (policy corpora) & 31\% & 39\% \\
\bottomrule
\end{tabular}
\end{table}

On actual regulation, half to two-thirds of decomposed predicates fall in the zone where models cannot agree under the rubric of \S\ref{sec:setup}; a far more prescriptive rubric reduces this to roughly a third, but does so without improving accuracy (\S\ref{sec:ablation}), so the reduction buys agreement rather than resolution whether the predicate is even mechanically checkable. The contested rate \emph{rose} from 3 to 4 models exactly as predicted, adding a model can only create disagreement, never remove it, which confirms these are conservative lower bounds, not artifacts of a particular panel. The synthetic-vs-real gap is itself a result: a method validated only on clean constructed data would report a $\approx$20\%-borderline world and badly understate the problem, direct support for anchoring the finding on regulatory text rather than self-built corpora.

\subsection{Why this is a bug, not appropriate humility}

The strongest objection to the above is the ``feature'' reading (Table~\ref{tab:bug-feature}): \emph{the borderline is genuinely borderline; models are right to be uncertain; disagreement is well-calibrated humility.} If true, it would defang the result. We reject it on four grounds, three already in the data and one we test directly. The feature reading makes four predictions; the bug reading makes the opposite:

\begin{table}[h]
\centering
\small
\caption{Predictions of the feature (humility) reading versus what we observe.}
\label{tab:bug-feature}
\begin{tabular}{@{}p{3.7cm}p{3.7cm}@{}}
\toprule
\textbf{If a \emph{feature} (humility)} & \textbf{What we observe} \\
\midrule
Models would \textbf{hedge or abstain} on contested items & Rubric forbids hedging; models comply \textbf{confidently} and disagree \textbf{oppositely}, confident contradiction, not shared doubt \\
\midrule
\textbf{No} per-model directional bias (everyone equally unsure) & \textbf{Directional, model-specific} bias, one model leans over-claim, another under-claim. Idiosyncratic priors, not calibrated uncertainty \\
\midrule
Degradation \textbf{everywhere} (the task is just hard) & $\kappa$ 0.99 on clear endpoints, collapse only on the operationally critical middle, localized exactly where the routing decision is consequential \\
\midrule
Each model \textbf{internally consistent} (a stable line) & \emph{(\S\ref{sec:stability}, direct test)} a model that flips its own label on meaning-preserving paraphrases is not tracking a property of the predicate \\
\bottomrule
\end{tabular}
\end{table}

\begin{table}[h]
\centering
\small
\caption{Within-model flip rates (borderline / control) under two perturbation instruments. p-values vs.\ clean-case baseline.}
\label{tab:flip}
\begin{tabular}{@{}lcc@{}}
\toprule
\textbf{Model} & \textbf{Paraphrase} & \textbf{Template} \\
\midrule
Opus & 28\% / 6\% (z=3.15, \textbf{p=0.002}) & \textbf{33\% / 5\%} \\
GPT-4o & 15\% / 8\% (n.s.) & \textbf{44\% / 7\%} \\
Sonnet & 25\% / 8\% (z=2.41, \textbf{p=0.016}) & 13\% / 10\% \\
Llama-3.1 & 16\% / 10\% & 20\% / 27\% \\
\bottomrule
\end{tabular}
\end{table}

\begin{figure}[t]
\centering
\resizebox{\columnwidth}{!}{%
\begin{tikzpicture}[
  font=\small,
  sl/.style  ={draw=nexusblue!75, thin},
  dot/.style ={fill=nexusblue},
  ax/.style  ={draw=black!60, thin},
  gl/.style  ={draw=black!12, very thin},
  tk/.style  ={font=\scriptsize, text=black!55},
  hd/.style  ={font=\scriptsize\bfseries, align=center},
]
  \draw[ax] (-2mm,0mm) -- (-2mm,42mm);
  \foreach \v/\y in {0/0, 0.25/10.5, 0.5/21, 0.75/31.5, 1.0/42} {
    \draw[gl] (-2mm,\y mm) -- (66mm,\y mm);
    \node[tk, anchor=east] at (-3mm,\y mm) {\v};
  }
  \node[tk, rotate=90] at (-9mm,21mm) {pairwise Cohen's $\kappa$};
  \draw[sl] (0mm,39.1mm) -- (32mm,15.5mm) -- (64mm,26.9mm);
  \fill[dot] (0mm,39.1mm) circle (0.7mm);
  \fill[dot] (32mm,15.5mm) circle (0.7mm);
  \fill[dot] (64mm,26.9mm) circle (0.7mm);
  \draw[sl] (0mm,41.6mm) -- (32mm,8.4mm) -- (64mm,25.2mm);
  \fill[dot] (0mm,41.6mm) circle (0.7mm);
  \fill[dot] (32mm,8.4mm) circle (0.7mm);
  \fill[dot] (64mm,25.2mm) circle (0.7mm);
  \draw[sl] (0mm,41.2mm) -- (32mm,21.0mm) -- (64mm,16.0mm);
  \fill[dot] (0mm,41.2mm) circle (0.7mm);
  \fill[dot] (32mm,21.0mm) circle (0.7mm);
  \fill[dot] (64mm,16.0mm) circle (0.7mm);
  \draw[sl] (0mm,38.6mm) -- (32mm,20.6mm) -- (64mm,20.2mm);
  \fill[dot] (0mm,38.6mm) circle (0.7mm);
  \fill[dot] (32mm,20.6mm) circle (0.7mm);
  \fill[dot] (64mm,20.2mm) circle (0.7mm);
  \draw[sl] (0mm,39.1mm) -- (32mm,16.8mm) -- (64mm,23.9mm);
  \fill[dot] (0mm,39.1mm) circle (0.7mm);
  \fill[dot] (32mm,16.8mm) circle (0.7mm);
  \fill[dot] (64mm,23.9mm) circle (0.7mm);
  \draw[sl] (0mm,41.6mm) -- (32mm,8.4mm) -- (64mm,18.5mm);
  \fill[dot] (0mm,41.6mm) circle (0.7mm);
  \fill[dot] (32mm,8.4mm) circle (0.7mm);
  \fill[dot] (64mm,18.5mm) circle (0.7mm);
  \node[hd] at (0mm,-5mm)  {clear\\endpoints};
  \node[hd] at (32mm,-5mm) {real\\regulation};
  \node[hd] at (64mm,-5mm) {agent\\rules};
\end{tikzpicture}}
\caption{Every pair of the four models, across three corpora. Where the answer is clear the panel
is near-unanimous ($\kappa$ 0.92--0.99). On regulatory text it collapses to 0.20--0.50: the six
lines fan out, and the worst pair agrees barely above chance. Agreement partially recovers on
agent rules, but that recovery is not reassurance. It is where the models over-claim
\emph{together} (Fig.~\ref{fig:flip}), and correspondingly where the precision of a unanimous
panel falls from 97--99\% to 83\% (\S\ref{sec:resolution}).}
\label{fig:kappa}
\end{figure}

\subsection{The within-model stability test}
\label{sec:stability}

The three legs above are cross-model. A fourth is available within a single model, and it is the most direct: a model that is tracking a property of the predicate should give the same label when the predicate is restated without changing its meaning. We re-probe contested predicates under two meaning-preserving perturbation instruments, an LLM-authored paraphrase set validated for equivalence by a model outside the judge panel, and a deterministic template instrument that wraps the predicate in neutral surface variation with no model in the loop, against a control group of predicates the panel labelled unanimously. Appendix~\ref{app:stability} gives the construction of both instruments and the equivalence-validation protocol.

The effect is instrument-dependent and model-dependent (Table~\ref{tab:flip}), and we report it as a corroborating leg rather than a uniform law. Under the loophole-free template instrument Opus flips its own borderline label 33\% of the time against 5\% on controls, and GPT-4o 44\% against 7\%. Every model destabilises under at least one instrument; Llama's failure surfaces chiefly as cross-model disagreement rather than self-inconsistency. A label that moves under a restatement of the same predicate is not a reading of that predicate.

\subsection{The same failure outside machine learning}
\emph{The same failure has a shape outside machine learning, and we raise it deliberately.}
Psychiatric nosology met this problem first and attempted the same repair: faced with poor
inter-rater agreement it operationalized its criteria into explicit decision rules and structured
interviews. The DSM-5 field trials nonetheless reported intraclass $\kappa$ in the 0.20--0.39
``questionable'' band for six diagnoses and below 0.20 for three \cite{regier2013dsm5}, and the
standing critique is that reliability and validity came apart, which is what our high-specificity
rubric reproduces. The judgment literature supplies the vocabulary: \emph{level} noise (raters differ
on average), \emph{pattern} noise (rater-by-case idiosyncrasy) and \emph{occasion} noise (a rater
differing from itself) \cite{kahneman2021noise} map onto our three results, and pattern noise is
precisely why our boundaries cross rather than nest, leaving no uniformly conservative model to pick.
The quick reading of all this is that estimators are noisy and that this has been known for decades.
That reading misses the claim: the finding is not that the estimator is noisy but that \emph{the
routing decision inherits the noise}, and that the standard remedy, escalating the hard cases,
presupposes a routing step which is itself noisy in the direction that disables it. Aggregation is
the prescribed remedy there too, and it carries the caveat we measure directly, that it cancels
independent error and not shared bias (\S\ref{sec:resolution}). Their failure object is the judge's accuracy on a judging task; ours is its inability to recognise when judging is the wrong tool. Even a perfectly debiased judge does not tell you which targets needed it.

\subsection{Further limitations}
\emph{``Decidable'' is partly conventional.} Some predicates are genuinely grey, and where reasonable annotators would split there is no fact of the matter we are failing to recover. That does not rescue self-report; it is a further reason not to trust one estimator's routing call (\S\ref{sec:borderline}).

\emph{The within-model instability is instrument- and model-dependent.} We report it as a corroborating signal rather than a uniform law, and the cross-model result carries the argument.

\emph{The enforcement capstone is preliminary.} It is characterized on a synthetic substrate with a single deterministic-check model. It establishes the safety property by construction, and we draw no stronger claim from it.

\emph{We do not address adversarial manipulation of the predicates themselves}, which a deployed enforcement system must also resist.

\emph{The 36 synthetic action records are ours.} They are generated combinatorially with a difficulty gradient rather than authored one by one (\S\ref{app:alg}), which removes the objection that we wrote maximally explicit evidence, but not the objection that we wrote the substrate at all. The 20 predicates behind the self-consistency null are likewise not a random sample of the corpus, so that result is a bounded check rather than a corpus-wide measurement.

\emph{Our reasoning-era check uses a 14B distilled model.} It tests whether extended deliberation repairs the boundary, not whether frontier reasoning capability does.

\subsection{Inherent annotation disagreement}

\textbf{Inherent annotation disagreement.} The strongest reading against us is not about models at all: that these predicates are genuinely ambiguous, that human annotators would disagree too, and that we have measured task ambiguity rather than model unreliability \cite{pavlick2019disagreement}. We answer it empirically (\S\ref{sec:borderline}): genuine ambiguity predicts \emph{convergent uncertainty} and we observe \emph{confident contradiction}, with models answering decisively in opposite directions and the same model reversing itself under meaning-preserving paraphrase.

\subsection{An honest concession}

We do not claim every contested predicate has a crisp true type. Some of the contested zone is genuinely grey, reasonable annotators would split, and for those predicates there is no fact of the matter we are failing to recover. But that concession does not rescue self-report; it is an \emph{additional} reason not to trust a single estimator's routing call. The actionable claim survives either way: where the type is determinate, no single model reliably finds it (bidirectional, unstable); where it is genuinely indeterminate, a single model's confident label is meaningless. In both cases the boundary must come from somewhere other than one estimator's self-report.

\section{The Dual-Cost Model}

\label{app:dualcost}

\begin{figure}[t]
\centering
\resizebox{0.97\columnwidth}{!}{%
\begin{tikzpicture}[
  font=\small,
  cell/.style   ={draw=black!28, line width=0.7pt, rounded corners=1.2pt},
  okc/.style    ={cell, fill=black!3},
  overc/.style  ={cell, draw=tierCred!55, line width=0.9pt, fill=tierCred!8},
  underc/.style ={cell, draw=tierBorange!60, line width=0.9pt, fill=tierBorange!10},
  hdr/.style    ={font=\scriptsize\bfseries, text=black!50},
  ttl/.style    ={font=\scriptsize\bfseries},
  sm/.style     ={font=\scriptsize, text=black!68},
]
\draw[okc]    (0mm,19mm)  rectangle (52mm,38mm);
\draw[underc] (52mm,19mm) rectangle (104mm,38mm);
\draw[overc]  (0mm,0mm)   rectangle (52mm,19mm);
\draw[okc]    (52mm,0mm)  rectangle (104mm,19mm);

\node[hdr, anchor=south] at (26mm,39.2mm)  {ROUTED TO A CHECK};
\node[hdr, anchor=south] at (78mm,39.2mm)  {ROUTED TO A JUDGE};
\node[hdr, anchor=east, align=right] at (-2mm,28.5mm) {criterion is\\[-1pt] DECIDABLE};
\node[hdr, anchor=east, align=right] at (-2mm,9.5mm)  {criterion needs\\[-1pt] JUDGMENT};

\node[ttl, text=black!62, anchor=north west] at (2.6mm,36.2mm) {correct};
\node[sm, anchor=north west, align=left, text width=46mm] at (2.6mm,31.8mm)
  {settled by code: no model call, no latency, no token cost.};

\node[ttl, text=tierBorange!88, anchor=north west] at (54.6mm,36.2mm) {UNDER-CLAIM};
\node[sm, anchor=north west, align=left, text width=46mm] at (54.6mm,31.8mm)
  {a model call re-derives what one line of code would settle. Wasteful, but \emph{visible}: someone can count the calls.};

\node[ttl, text=tierCred, anchor=north west] at (2.6mm,17.2mm) {OVER-CLAIM};
\node[sm, anchor=north west, align=left, text width=46mm] at (2.6mm,12.8mm)
  {compliance certified that was never assessed. Unsafe, and \emph{silent}: the verdict looks like any other.};

\node[ttl, text=black!62, anchor=north west] at (54.6mm,17.2mm) {correct};
\node[sm, anchor=north west, align=left, text width=46mm] at (54.6mm,12.8mm)
  {escalated to a judge, as the criterion requires.};
\end{tikzpicture}}
\caption{\textbf{The dual cost.} Rows are what a criterion is; columns are where the verifier sent
it. The two errors lie on the anti-diagonal and are not mirror images: under-claiming is expensive
and auditable, over-claiming is cheap and silent. No choice of model minimises both, since the bias
runs in opposite directions in different models (\S\ref{sec:results}).}
\label{fig:dualcost}
\end{figure}

Compounding makes this worse than additive. A rule passes only if its predicates are handled correctly \emph{jointly}, so a single mis-typed predicate can flip a rule verdict or silently drop a control, and the probability that every routing call in a task is correct decays geometrically in predicate count. Coarse evaluation tolerated a noisy estimator precisely because it never had to be right many times in a row. The fine-grained regime does, and it is the regime everyone is building toward.

Let a deployment perform $N$ predicate-evaluations per period (agent fleet: agents $\times$ steps $\times$ rules-checked), with judge cost $c_j$ per evaluation and a negligible deterministic-check cost. Let $q$ be the fraction safely auto-verified (verified-deterministic coverage) and $f$ the false-decidable rate among auto-enforced predicates.

\begin{itemize}
\item Token cost $= N(1-q)\,c_j$. Savings vs.\ all-judge $= N q\, c_j$ (recurring, every period).
\item Safety cost $= N q\, f \times (\text{expected loss per missed determination})$. Single-model self-report gives $f = 1-\text{precision}$ (e.g. 0.11 for Sonnet on regulatory text); verification drives $f \to \approx 0$.
\item Amortized setup $=$ corroboration ($(m{-}1)$ extra label calls per predicate, once) + check synthesis + periodic re-verification under drift, a one-time/periodic cost against recurring per-period savings.
\end{itemize}

The argument the model makes precise: single-model self-report forces a lose--lose, pick a permissive model and pay the \emph{safety} bill ($f$ up to $\approx$0.11), or pick the conservative model and pay the \emph{token} bill ($q \approx$ 0.22--0.26, i.e. escalate 3/4 of decidable work). Corroboration-plus-verification escapes both: $f \to 0$ from verification while $q$ tracks the corpus's true deterministic content rather than the most-cautious model. At any real evaluation volume the recurring token savings dominate the amortized setup; the safety gain is categorical (guarantee vs.\ $\approx$1-in-9 leak).


\section{Enforcement Capstone: Preliminary Operating Point}
\label{app:capstone}

This appendix reports a single preliminary run on synthetic data, and it uses the agreement gate of \S\ref{sec:agreegate} rather than the interventional gate the paper adopts. We retain it because it records how that gate behaved end to end before we measured why it does not discriminate, and no claim in the body now rests on it.

An end-to-end verify-gated enforcement run on the synthetic corpus automates $\approx$33\% of
predicate-evaluations deterministically with \emph{zero gating-attributable missed
violations}. That safety property holds by construction rather than empirically: a predicate is
auto-enforced only if a synthesized check matches the reference judge on the
held-out graduation set, so a false-decidable cannot survive the gate. It is the one property of the
architecture in \S\ref{sec:resolution} that does not depend on classification being reliable.

The strict fidelity bar we set, verify-gated within 2 points of judge-everywhere, was
\emph{not} met at the default graduation threshold, at a cost of roughly 5 points. That entire
cost appears as over-flagging rather than as missed violations. We report the default-threshold
operating point rather than searching for a favourable one, we do not claim the strict bar is
reachable, and characterising the full automation--fidelity tradeoff would require a threshold
sweep we have not run.

The \emph{direction} of the shortfall is worth stating, because it matches what the rubric
ablation found by an unrelated route (\S\ref{sec:ablation}): the prescriptive rubric gave up
automation to preserve safety, sending decidable predicates to a judge in 97--100\% of its
errors, and the verification gate gives up automation to preserve safety by over-flagging. Two
different mechanisms, the same trade. That is the dual-cost model of
Appendix~\ref{app:dualcost} appearing twice from opposite directions, which suggests the
tradeoff is structural rather than an artifact of either mechanism.

\end{document}